\documentclass{bmvc2k}

\usepackage{multicol,multirow}
\usepackage{placeins}
\usepackage{tipa}
\usepackage{amsmath}
\usepackage{amssymb,bm}

\title{Visual speech recognition:\\ aligning terminologies for better understanding}

\addauthor{Helen L Bear}{helen@uel.ac.uk}{1}
\addauthor{Sarah L Taylor}{s.l.taylor@uea.ac.uk}{2}

\addinstitution{
Computer science and informatics\\
University of East London\\
 London E16 2RD UK
}
\addinstitution{
School of computing sciences\\
University of East Anglia\\
 Norwich NR4 7TJ UK
}

\runninghead{Bear, Taylor}{Visual speech recognition: A mini review}

\def\eg{\emph{e.g}\bmvaOneDot}

\def\etal{\emph{et al}\bmvaOneDot}

\begin{document}
\maketitle
\begin{abstract}
We are at an exciting time for machine lipreading. Traditional research stemmed from the adaptation of audio recognition systems. But now, the computer vision community is also participating. This joining of two previously disparate areas with different perspectives on computer lipreading is creating opportunities for collaborations, but in doing so the literature is experiencing challenges in knowledge sharing due to multiple uses of terms and phrases and the range of methods for scoring results. 

In particular we highlight three areas with the intention to improve communication between those researching lipreading; the effects of interchanging between speech reading and lipreading; speaker dependence across train, validation, and test splits; and the use of accuracy, correctness, errors, and varying units (phonemes, visemes, words, and sentences) to measure system performance. We make recommendations as to how we can be more consistent.

\end{abstract}

\section{Introduction}

Machine lipreading is the classification of speech from only visual cues of a speaker and has long been a niche research problem in the field of speech processing. However, recent developments in deep learning has attracted significant interest and advancements from the Computer Vision and Machine Learning communities.


The first machine lipreading approaches were adaptations of conventional audio-based speech recognition systems. Some of the challenges of machine lipreading are common to acoustic recognition but require different approaches (\eg modelling coarticulation), whereas some are unique to the visual channel (\eg normalising for environmental lighting and speaker pose). It is reasonable to assert that both speech processing and computer vision techniques are complementary in solving the issues associated with machine lipreading. However, it is important that, as we bridge the gap between the two areas of research, both terminology and performance metrics are somewhat standardised. 



In this paper, we address some potential discrepencies in terminology, experimental setup and performance reporting that appear in both speech and vision publications. The structure of this paper is as follows; first we clarify the distinction between \emph{speech reading} and \emph{lipreading}. We then discuss the challenges of speaker dependence in lipreading systems, and define the scope of this problem. Next we summarise the different metrics currently used to report the performance of lipreading machines. Finally, we suggest some recommendations for moving forward with the same understanding of certain terms and how we can compare our scoring methods.


\section{Speech reading Vs Lipreading}
\label{sec:srvlr}
Despite commonly being used interchangeably \cite{petajan1984automatic}, the terms speech reading and lipreading have subtle but distinctive definitions. 

\begin{figure}[!ht]
\centering
\setlength\tabcolsep{1.5pt}
\begin{tabular}{cccccc}
\includegraphics[width=0.151\linewidth]{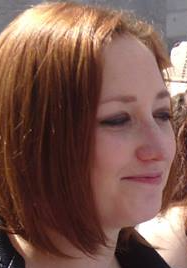} &
\includegraphics[width=0.155\linewidth]{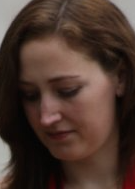} &
\includegraphics[width=0.16\linewidth]{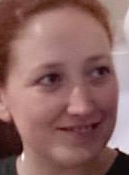} &
\includegraphics[width=0.151\linewidth]{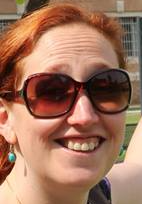}&
\includegraphics[width=0.143\linewidth]{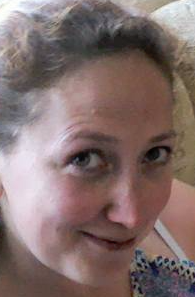} &
\includegraphics[width=0.164\linewidth]{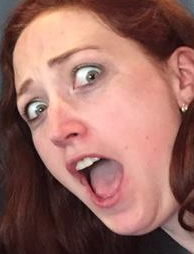} \\
\end{tabular}
\caption{Regions of an image used in speech reading.}
\label{figspeechRead}
\end{figure}

\textbf{Speech reading} is what human lip readers do. They interpret speech using information provided by the whole face and body since knowledge of the facial expression, gaze and body gestures often helps to provides semantic context that makes decoding the speech easier. In the computer science domain, machine speechreading systems usually use just the face to decode the speech content. Figure~\ref{figspeechRead} contains examples of the region of an image that might be used for machine speech reading.

\begin{figure}[!ht]
\centering
\centering
\setlength\tabcolsep{1.5pt}
\begin{tabular}{cccccc}
\includegraphics[width=0.15\linewidth]{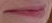} &
\includegraphics[width=0.15\linewidth]{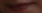} &
\includegraphics[width=0.15\linewidth]{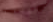} &
\includegraphics[width=0.15\linewidth]{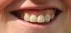}&
\includegraphics[width=0.15\linewidth]{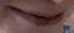} &
\includegraphics[width=0.15\linewidth]{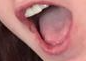} \\
\end{tabular}
\caption{Regions of an image used in lipreading.}
\label{figlipRead}
\end{figure}

\textbf{Lipreading} is the interpretation of speech from the motion of the lips alone (see Figure~\ref{figlipRead}). This is the region of the image considered by the early machine lipreading approaches, and does not contain any information regarding the upper facial expression or body language.  
%

Human lipreaders do not read lips, but in fact read cues provided by the speaker's entire body. Most people, including those with perfect hearing, use visual information from the speaker's face and body to decode speech when it is available. In some cases, for example in a noisy public house, a lipreader may focus on the lips of the speaker to prioritise cues from the lips over other information, but this does not discount the value of other data. Rather it focuses ones attention to where is it most useful. 

\begin{figure}[!ht]
\centering
\includegraphics[width=0.5\linewidth]{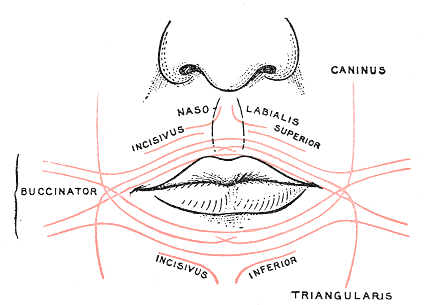}
\caption{Illustrating the connecting fibers from the orbicularis oris muscle that surrounds the lips \cite{lipmuscles}.}
\label{fig:lipmusclefibers}
\end{figure}


It is difficult to track the lips of a speaker in isolation. The lips have no skeletal structure and a deformable surface. The orbicularis oris muscle encircles the mouth and allows for lip puckering and protrusion and also plays a role in lip closure. It is composed of four interlacing independent quadrants and gives an appearance of circularity \cite{lipform}. The fibers of the orbicularis oris connect to other parts of the face. In Figure~\ref{fig:lipmusclefibers} we see that fibers from some of the cheek muscles, decussate (form an `X' shape) around the lips, and strongly control lip motion. The connection of the fibers from the chin and nose have a smaller, but significant influence on lip motion.

\begin{figure}[!ht]
\centering
\begin{tabular}{c c}
\includegraphics[width=0.3\linewidth]{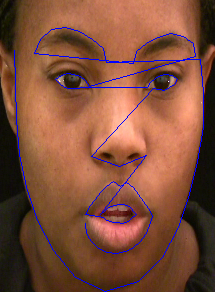} & 
\includegraphics[width=0.34\linewidth]{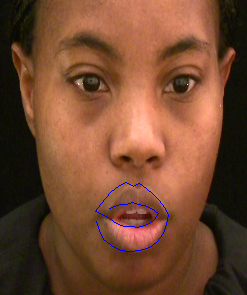} \\
\end{tabular}
\caption{Examples of shapes for a full face tracking (left), and a lip only track (right). We can see how the lip shape is mis-aligned with the actual lip shape}
\label{fig:fullfacelandmarks}
\end{figure}

We often track a whole face rather than only the lips throughout a video (Figure~\ref{fig:fullfacelandmarks}). The extra structure from the rest of the face enables easier tracking and a more robust fitting. For example, using active appearance models (AAMs) \cite{Matthews_Baker_2004} we achieve better lip tracking if we track the contour of the face and facial features in addition to the lips. 

Due to the informal use of the terms used in literature, and the use of the full face or lips only at different stages of the lipreading process, it is important for researchers to explicitly clarify whether the features that they are using are derived from the full face or from the lips. Not only is this necessary for reproducibility, but it also enables us to gauge the benefit from each approach.



\FloatBarrier
\section{Speaker independence}

First we clarify what we mean by speaker dependence and independence in lipreading. Speaker independence in machine lipreading is achieved when classification models generalise to spoken utterances by talkers not contained within the training set. If a system only works on a closed set of speakers, or is not tested on speakers that are outside of the training set, we can assume that the approach is speaker dependent.

To explain with examples, dataset A contains $1,000$ utterances by speaker $X$. To build a speaker dependent lipreading system, we can use, for example, $800$ utterances as training samples and $200$ as test samples. To achieve speaker independence, it is not sufficient to only separate specific utterances of a speaker, i.e. sentences $1$ to $n$ for speaker $N$ to train, and sentences $n+1$  to $1,000$ for test. 

\begin{figure}[!ht]
\centering
\resizebox{\columnwidth}{!}{%
\begin{tabular}{c c c || c}
\multicolumn{3}{c||}{Training speakers} & Test speakers \\
\includegraphics[width=0.223\textwidth]{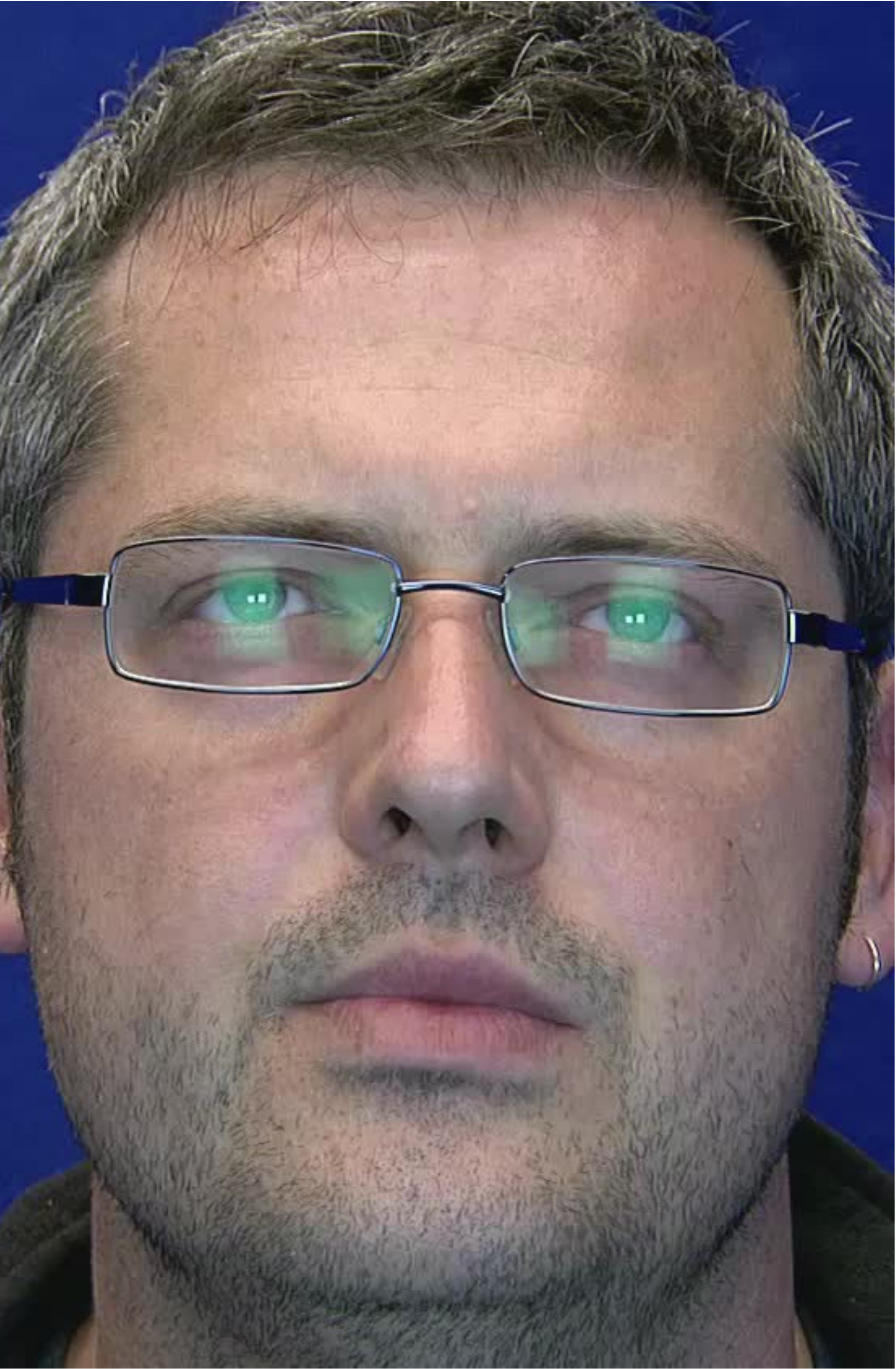} &
\includegraphics[width=0.25\textwidth]{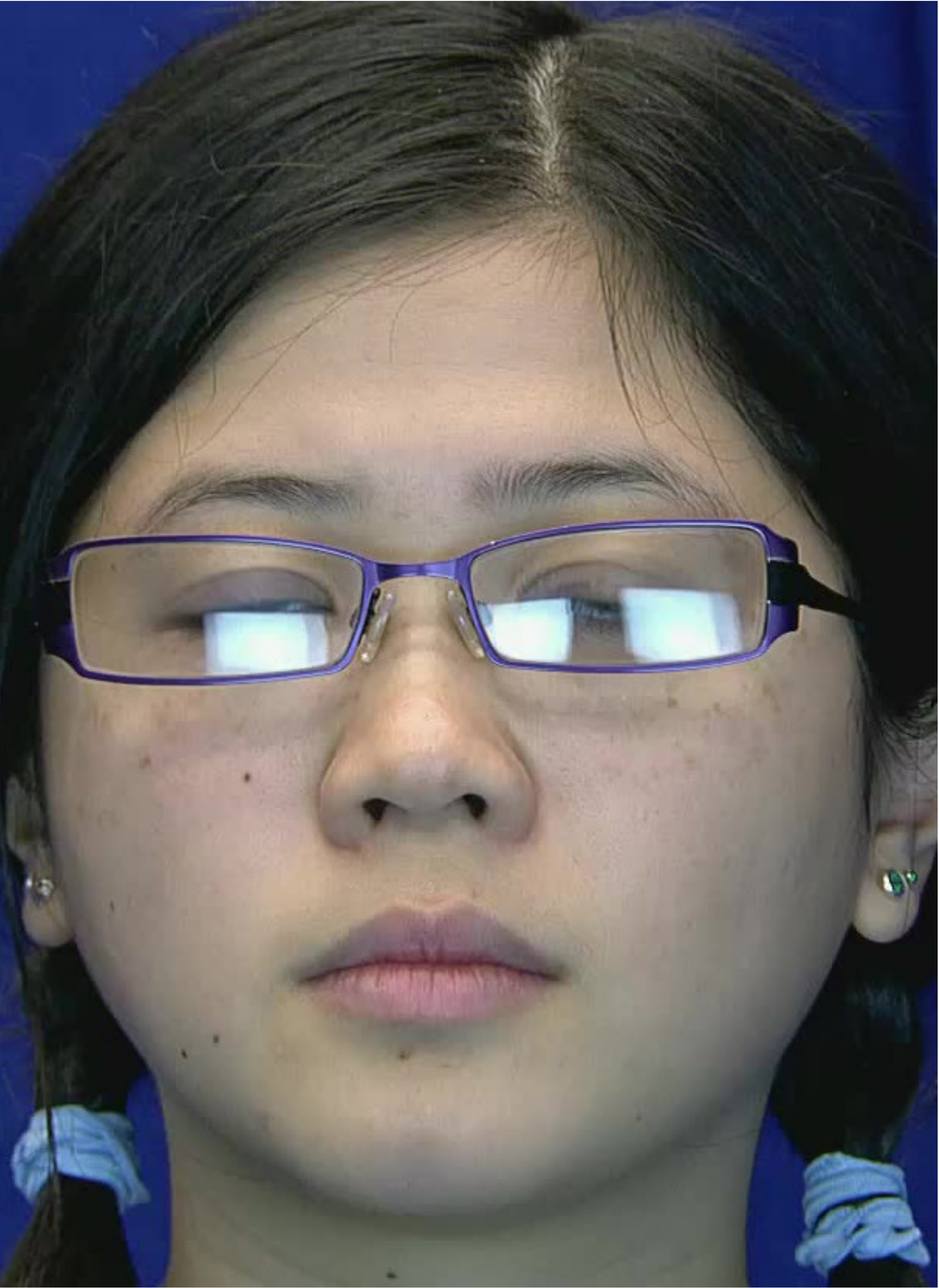} &
\includegraphics[width=0.255\textwidth]{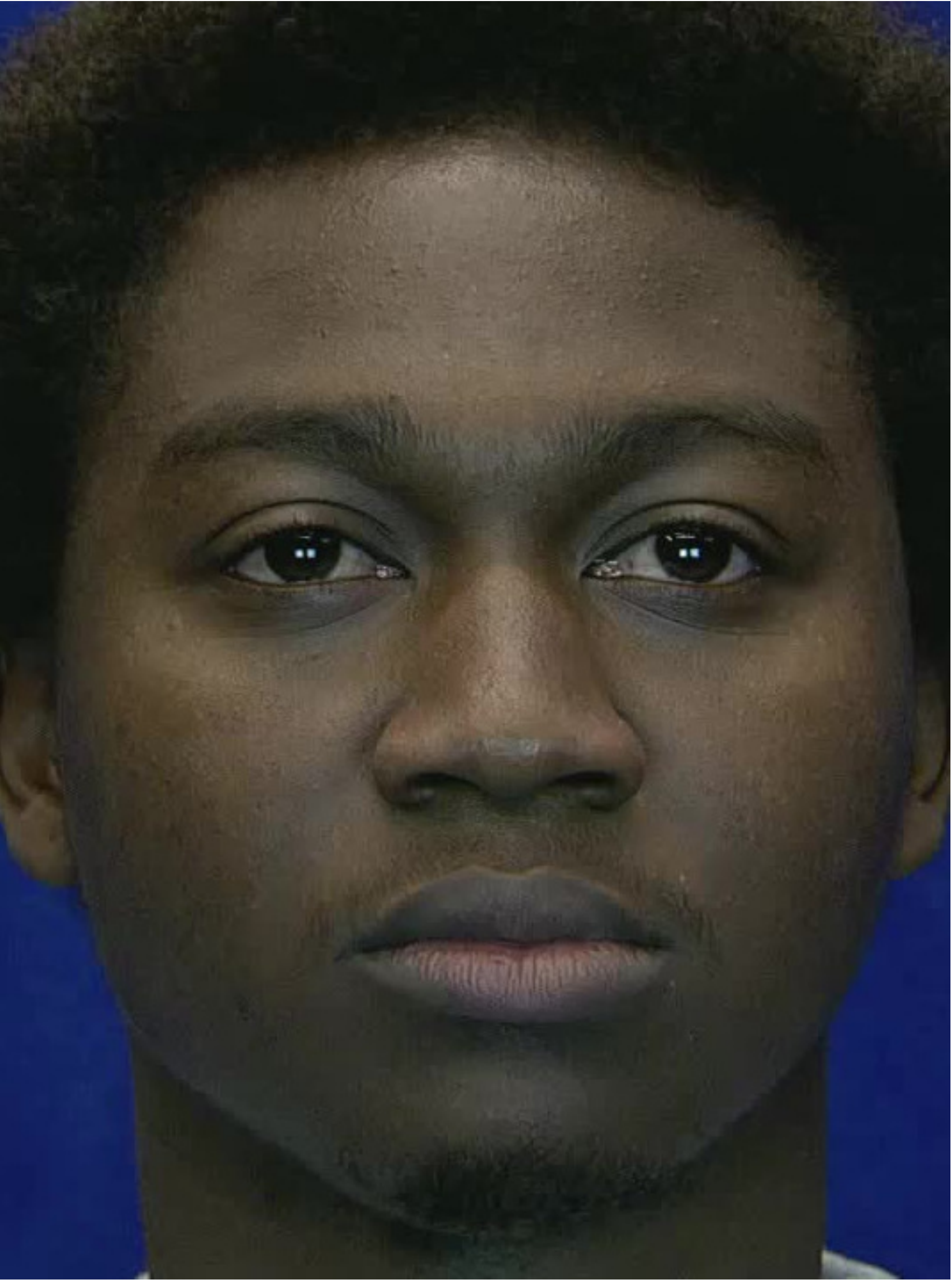} &
\includegraphics[width=0.275\textwidth]{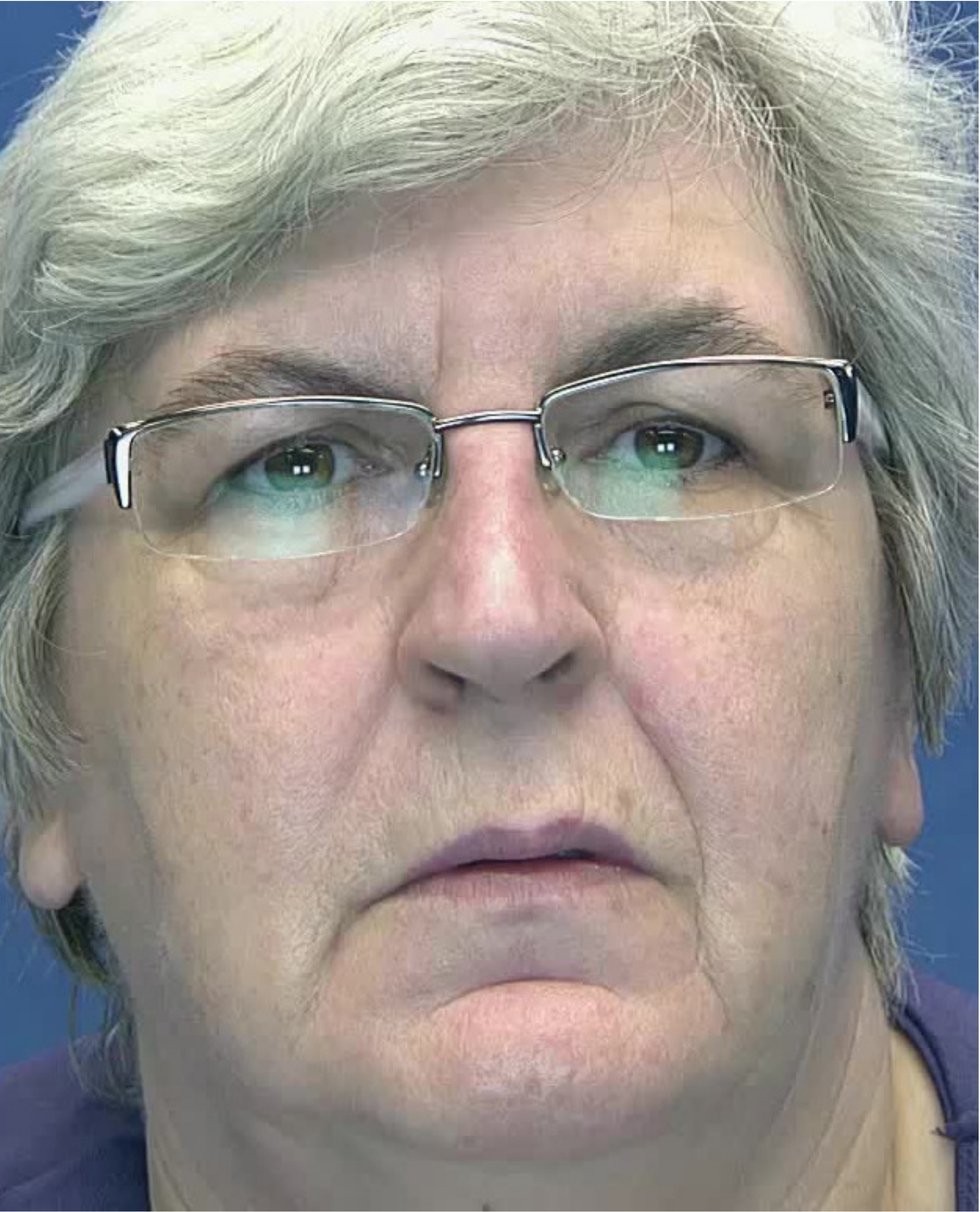} \\
$Sp_w$ & $Sp_x$ & $Sp_y$ & $Sp_z$ \\
\end{tabular}%
}
\caption{Speaker independence in data divisions.}
\label{fig:si}
\end{figure}

Alternatively, if we have speaker $X$ and speaker $Y$ in a dataset, each with $1,000$ utterances, we can use the $800$ speaker $X$ training samples to build our classifier, and we would test on the $200$ samples of speaker $Y$ and vice versa. This is speaker independent lipreading, see Figure~\ref{fig:si}. It ensures that the model is learning a classifier that is not biased by the identity of the speaker, and is generalisable to new speakers.

\begin{figure}[!ht]
\centering
\resizebox{\columnwidth}{!}{%
\begin{tabular}{c c || c c c || c}

\multicolumn{2}{c||}{Training speakers} & \multicolumn{3}{c||}{Validation speakers} & Test speakers \\
\includegraphics[width=0.2\textwidth]{figs/s2} &
\includegraphics[width=0.202\textwidth]{figs/s3} &
\includegraphics[width=0.2\textwidth]{figs/s2} &
\includegraphics[width=0.202\textwidth]{figs/s3} &
\includegraphics[width=0.192\textwidth]{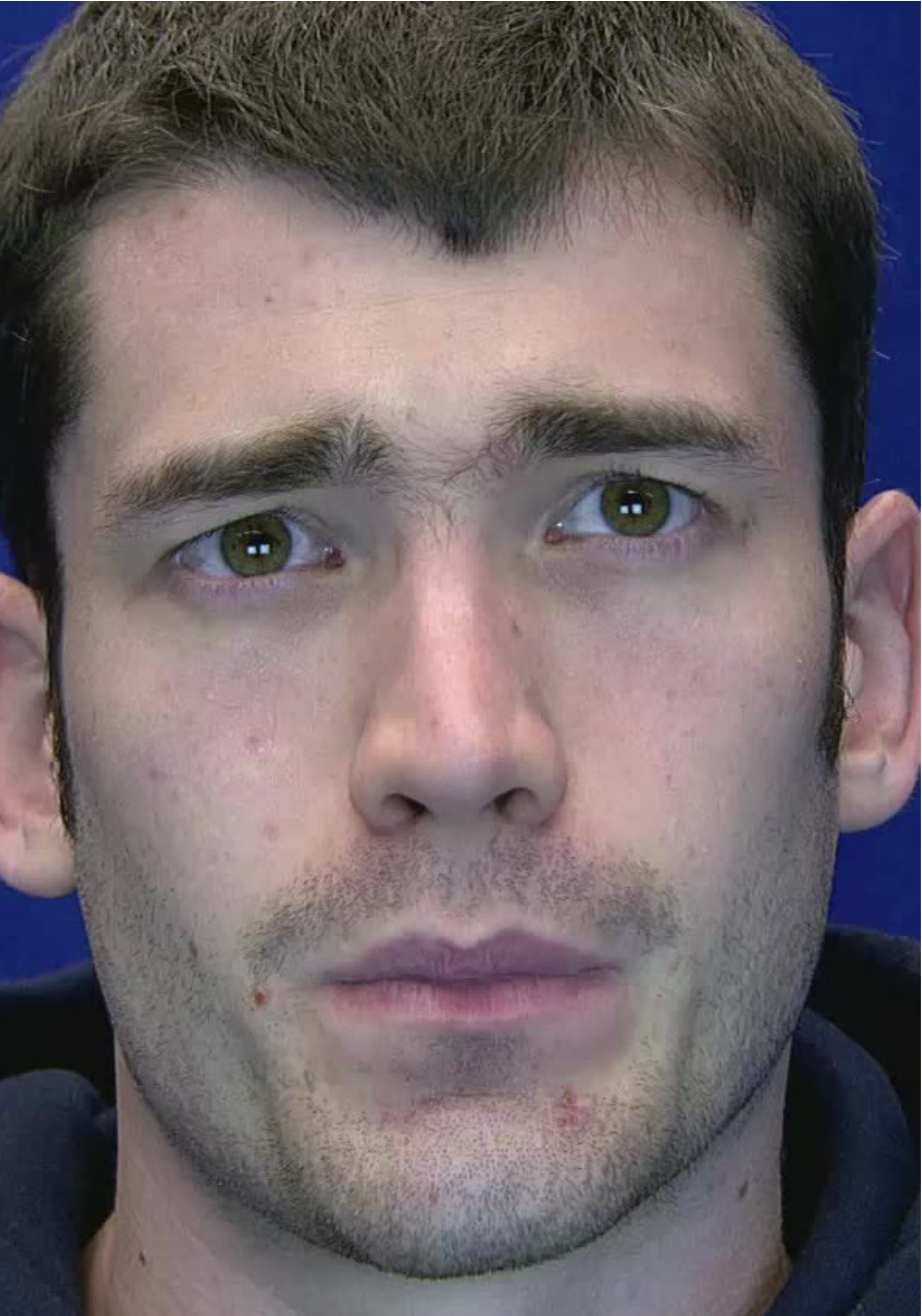} &
\includegraphics[width=0.22\textwidth]{figs/s4}
\\
$Sp_x$ & $Sp_y$ & $Sp_x$ & $Sp_y$ & $Sp_a$ & $Sp_z$ \\
\end{tabular}%
}
\caption{Speaker independence in data divisions.}
\label{fig:sd}
\end{figure}

For classification methods that require the data to be divided into train, validation, and test sets, the validation set can contain speakers from the training set and new speakers, but speakers must remain distinct from the test set if speaker independence is the goal. See Figure~\ref{fig:sd}. Note that for any duplicate speakers in both training and validation sets, one must split samples between the two, i.e. sample $1$ for speaker $Sp_x$ can only be in either training or validation. 


Traditionally, machine lipreading systems were speaker dependent, to the degree that a separate model was built for each speaker. Modern approaches exploit extremely large datasets to train speaker independent models~\cite{chung2016lip} with good performance. Sadly, these datasets are not publically available. 


The challenge of speaker independent machine lipreading was discussed by Cox \etal \cite{cox2008challenge}, they concluded that with multispeaker classifiers the accuracy degrades significantly and offered two, not exclusive, explanations;
\begin{itemize}
\item current features are not good enough. Better features would encode the information in an utterance independently of the physiology of the speaker and what he or she does with their mouth when they speak.
\item that there is more inherent variability in lipreading than in acoustic speech recognition.
\end{itemize}

Anecdotal evidence for human lipreaders supports this second option who suggest that it takes time to adjust to new talkers, and it is easier to lip read those one already knows. 

We can but hope for new generalisation methods and large, freely available audio-visual datasets for this to be addressed, but in the meantime, studies have attempted to measure the influence of speaker dependence, for example, \cite{bear2015speakerindep} and \cite{bear2017bmvcIndep}. Understanding the influence of speaker identity in the speaker independence problem, will enable researchers to tackle the speaker adaptation question, but a full assessment of speaker identity is beyond the scope of this review. 

One possible explanation as to why speaker identity is a significant affect on lipreading is the implementation of the conventional tracking methods. When tracking faces through videos, we have a choice of:
\begin{enumerate}
\item one tracking model, trained on labeled samples from all speakers in a dataset, or,
\item a set of tracking models, each one only trained on samples from each speaker individually.
\end{enumerate}  
In practice, option two achieves the most accurate fit throughout a video and this ensures that robust features are extracted.

However, the features become speaker specific, particularly for models that encapsulate appearance information, meaning that the speaker specificity is \emph{ingrained} into the features, thus making speaker generalisation more challenging for training any classification model. 

\begin{table}[!ht]
\centering
\caption{Example feature vector sizes}
\resizebox{\columnwidth}{!}{%
\begin{tabular}{|l|c|c|c|c|c|c|c|c|c|c|c|c|}
\hline
Speaker 	& 1  & 2  & 3  & 4  & 5  & 6  & 7  & 8  & 9  & 10 & 11 & 12 \\
\hline
Shape 		& 13 & 13 & 13 & 13 & 13 & 13 & 13 & 13 & 13 & 14 & 13 & 13 \\
Appearance 	& 46 & 47 & 43 & 46 & 45 & 45 & 47 & 45 & 47 & 72 & 45 & 37\\
\hline
Total 		& 59 & 60 & 56 & 59 & 58 & 58 & 60 & 58 & 60 & 86 & 58 & 50 \\
\hline
\end{tabular}%
}
\label{tab:featureSizes}
\end{table}

As an example, in Table~\ref{tab:featureSizes} we show the size of the AAM feature vectors representing the tracked lip region of 12 speakers from the RMAV audio-visual dataset \cite{lan2010improving} split by shape and appearance. Whilst the shape of the lips for each speaker can be represented by a remarkably consistent number of features, the number of appearance features vary from $43$ to $72$. This means that different faces require appearance models of variable complexity to accurately represent their particular speech movements, and generalisation is difficult.

\FloatBarrier

\section{Scoring metrics and methods}
Methods of reporting on the performance of machine lipreading have been adopted from audio speech recognition systems. The most common published figures are correctness and accuracy as shown in Equations~\ref{eq:corr} and~\ref{eq:acc} respectively \cite{young2006htk}. 

\begin{equation}
\centering
C=\frac{N-D-S}{N} \qquad or \qquad C\%=\frac{N-H}{N} \times 100\%
\label{eq:corr}
\end{equation}
\begin{equation}
\centering
A=\frac{N-D-S-I}{N} \qquad or \qquad A\%=\frac{N-H-I}{N} \times 100\%
\label{eq:acc}
\end{equation}
Where $N$ is the total number of labels in the ground truth, $D$ is the number of deletion errors, $S$ represents the number of substitution errors, and $I$ is the number of insertion errors after comparing a ground truth transcript to a recognition transcript. $H$ is the sum of $D+S$.

Conversely, error rates are also presented. In essence, these are the inverse of correctness and accuracy, see Equation~\ref{eq:wer}. 

\begin{equation}
\centering
ER\% = \frac{D - S - I}{N} \times 100\%
\label{eq:wer}
\end{equation}

However, these metrics can be used in two different ways in machine lipreading; firstly, by measuring the performance of the classifier output, labeled by the classifier unit (options are visemes, phonemes, or words), or secondly, by measuring the performance of the system after using a dictionary or language decoder (unit options at this stage of the lip reading system are visemes, phonemes, words, or sentences). We address these two variations in turn.

\subsection{Error types}
In speech we distinguish between type of error as these have a meaningful impact on interpretation. There is a difference in an estimated output being grammatically correct or simply understandable. Any threshold of lip reading performance will depend upon its purpose. For general conversation in the deaf community, understanding intent in a communicator's speech is prioritised over a precise transcription. However, for security settings, or evidence gathering, exact and confident transcriptions are essential to remove any ambiguity of what a speaker uttered.

We can explain the types of error shown in Equations~\ref{eq:acc} and~\ref{eq:wer} with an example. Suppose we have a ground truth utterance, ``John wanted to visit the shop to buy groceries". Our classifiers can produce different outputs. 

\textbf{Deletion errors}. Possible output 1: `` John wanted visit the to groceries" has three words missing; `to', `shop', and `buy'. In this instance, these are deletion errors. 
\textbf{Insertion errors}. In another possible output: ``John wanted to visit visit the shop to buy groceries'', the word `visit' is included twice. This is an insertion error. 
\textbf{Substitution errors}. Finally, if we achieved a classifier output of ``John wanted to shop the shop to buy groceries". The word `shop' has been identified where the word `visit' should be. This is a substitution error.

However, whilst we make this distinction between the types of error we encounter, it is standard practice to weight the influence of each error uniformly. A study by Satki \etal~\cite{7041620} instead weights them based upon brain signals (EPGs) during visual speech perception. Satki \etal's Equation~2 is a weighted word error rate (shown here in Equation~\ref{eq:weighted}), where the weights are calculated by time periods recorded on EPG error effects.
\begin{equation}
\centering
ERP-WER = \frac{\alpha D - \beta S - \lambda I}{N} * 100
\label{eq:weighted}
\end{equation}

They report that the insertion errors have the greatest negative influence on understanding, and suggest a weight of $\lambda = 0.3$ for these errors. We suggest this is because an insertion error is a distraction from the intended message. Deletion errors are assigned a weight of $\beta= 1.2$. We suggest this error is weighted higher because, although data is absent, the context offered by the preceding and succeeding labels enables understanding though lexical interpolation and prior language knowledge. Substitutions were weighted by $\alpha = 1.5$. One hypothesis for this is that there is a low difference between some visual classes. This means that class substitutions in an output transcript are likely to be the closely related classes and mislead an interpreter.

In machine lipreading, rather than human perception in \cite{7041620}, we are yet to measure the influence of each type of error on machine language decoding. 

\subsection{Metric units}
In machine lipreading, one attempts to interpret words spoken from the visual representation of sounds as uttered. This means there are two levels of `translation' within the process, visual gestures (known as visemes) into phonemes, and phonemes into words. Previous literature reports lipreading performance based on different units which makes comparing performance difficult. Some report word error rate \cite{lucey2009visual,4064511}, others viseme error rate \cite{bearicip}, or others accuracy of these two units \cite{kwanchivabmvc}. Some attempts have been made to compare phonemes to visemes \cite{cappelletta2012phoneme, kwanchivabmvc} and boost between them \eg \cite{bear2015findingphonemes} but as visemes are yet to be formally defined or understood (most researchers use working definitions \eg \cite{hilder2010pursuit}) we omit neither. 

It is possible, for example, to build a word classifier followed by a language model measured in terms of its viseme correctness. Such a system would be bizarre but is none-the-less possible. Table~\ref{tab:sn_tests} shows some of the more sensible possibilities. 

\begin{table}[h]
\centering
\caption{Unit selection pairs for classifiers \& language networks.} 
\begin{tabular}{|l|l|}
\hline
Classifier units & Language units  \\
\hline \hline
Viseme & Viseme \\
Viseme & Phoneme \\
Viseme & Word  \\
Viseme & Sentence \\
Phoneme & Phoneme \\
Phoneme & Word  \\
Phoneme & Sentence \\
Word & Word  \\
Word & Sentence \\
\hline
\end{tabular}
\label{tab:sn_tests}
\end{table}

\subsection{Notation suggestion}
We are not in a position to definitively select the `right' units, nor dictate the `best' metric to use as that is a choice for each researcher to decide. We can however make a simple recommendation which would help us to quickly and easily compare results. 

\begin{equation}
\centering
M_{cu} \qquad || \qquad  M^{nu}
\label{eq:newNotation}
\end{equation}
where $M$ is the metric (e.g. $A$, $C$, or $ER$), and subscript notation for a classifier output, or a superscript notation for a network/dictionary output. The subscript $cu$ could be any of the left column of Table~\ref{tab:sn_tests} \{v, p, w\}, and superscript $nu$ could be any of the right column of Table~\ref{tab:sn_tests} \{v, p, w, s\}. For example, $A_p$ would represent the accuracy of phoneme classifiers, and $ER^v$ would represent viseme error rate using a viseme based language model. 

It would be informative to report the scores achieved at both stages of the lipreading system to understand whether recognition performance is dependent on the language decoding step or by well-trained classifiers as this will vary between different lipreading system architectures. 

\FloatBarrier
\subsection{Performance evaluation}
It is often useful to visualise and report results from a confusion matrix, where we count the number of times a unit is recognised or confused with another (see Figure~\ref{table:examplecm}). When reading values from a confusion matrix we have choices. Either we can count the probability of class $\mbox{Pr}\{c|\hat{c}\}$ (if you look at the matrix as row major) or the inverse probability $\mbox{Pr}\{\hat{c}|c\}$ (if you view the matrix as column major), where $c$ represents a single class. 

\begin{figure}[!ht]
\centering
\resizebox{\columnwidth}{!}{%
\begin{tabular}{|l|c|c|c|c|c|c|c|c|c|c|c|c|c|c|c|}
\hline
& \multicolumn{15}{|c|}{Predicted classes} \\
\hline
\multirow{ 15}{*}{Actual Classes} & 76 & 2  & 1  & 5   & 2  & 1  & 3  & 1  & 1  & 3   & 1  & 5  & 0  & 0  & 4 \\ \cline{2-16}
&  0  & 28 & 0  & 1   & 0  & 2  & 0  & 0  & 1  & 0   & 0  & 0  & 0  & 0  & 2  \\ \cline{2-16}
&  0  & 4  & 17 & 0   & 1  & 2  & 0  & 0  & 1  & 0   & 0  & 1  & 0  & 0  & 0  \\ \cline{2-16}
&  3  & 6  & 6  & 163 & 3  & 7  & 7  & 2  & 8  & 7   & 1  & 4  & 2  & 0  & 1  \\ \cline{2-16}
&  4  & 2  & 2  & 3   & 33 & 0  & 0  & 1  & 3  & 0   & 1  & 2  & 1  & 0  & 1  \\ \cline{2-16}
&  2  & 0  & 0  & 6   & 1  & 9  & 0  & 0  & 1  & 0   & 0  & 0  & 0  & 0  & 0  \\ \cline{2-16}
&  4  & 1  & 0  & 3   & 1  & 1  & 40 & 0  & 0  & 1   & 5  & 2  & 0  & 0  & 0  \\ \cline{2-16}
&  0  & 3  & 2  & 1   & 2  & 0  & 0  & 11 & 8  & 2   & 0  & 2  & 0  & 0  & 1  \\ \cline{2-16}
&  0  & 0  & 1  & 4   & 1  & 0  & 1  & 2  & 97 & 3   & 1  & 0  & 0  & 0  & 0  \\ \cline{2-16}
&  2  & 1  & 4  & 1   & 2  & 3  & 0  & 1  & 6  & 110 & 8  & 0  & 2  & 0  & 0  \\ \cline{2-16}
&  0  & 1  & 0  & 1   & 0  & 1  & 2  & 0  & 0  & 1   & 14 & 1  & 2  & 0  & 0  \\ \cline{2-16}
&  0  & 0  & 0  & 3   & 1  & 1  & 4  & 1  & 1  & 3   & 1  & 16 & 1  & 0  & 0  \\ \cline{2-16}
&  0  & 0  & 0  & 0   & 0  & 0  & 0  & 0  & 0  & 0   & 1  & 0  & 3  & 0  & 0  \\ \cline{2-16}
&  0  & 0  & 0  & 0   & 0  & 0  & 0  & 0  & 0  & 0   & 0  & 0  & 0  & 84 & 0  \\ \cline{2-16}
&  1  & 3  & 0  & 2   & 1  & 1  & 1  & 0  & 0  & 1   & 2  & 0  & 0  & 0  & 28 \\  
\hline
\end{tabular}%
}
\caption{An example confusion matrix}
\label{table:examplecm}
\end{figure}

These can be described as two different questions, where the former is `I am looking for class $x$, is it $X$?' or the latter is `I have this, which class is it?'. Whilst most researchers use the former as this yields higher accuracy scores, when investigating inter-class variabilities, such as in \cite{bear2014some} it can be useful to use the inverse probabilities. As it is important to ensure understanding from the classifier output, we recommend that authors are clear which values their accuracy scores are calculated from for fair benchmarking. 

\FloatBarrier
\subsection{The top five/top one question}
It is not uncommon in computer vision literature to report top five values as measure of classification performance \eg \cite{NIPS2012_4824}. However, the inter-class variation in a set of phonemes or visemes is much smaller than, for example, that within a set of images representing, say cats, dogs, building, or landscape. If a cat and a dog image are confused, this is more reasonable than a confusion between a cat and a building.  However, the issue this causes in speech, is that the classified output transcript can have a significantly different meaning resulting in confusion between talkers. Imagine if one was saying `Pass me the salt' but due to misclassification in the top five classes,  we transcribed `Pass me the malt'. It is only one phoneme different (a $/m/$ instead of an $/s/$) but this single substitution is significant. Thus, speech researchers, as a general rule, only report the top one result as this means the original intention has been recognised accurately.

Thus we recommend that if the top five values are to be reported, then the top one value should be included also. We note that examples of the top five class confusions might be useful to understand relationships between certain classes, whether phonemes or visemes. 

\FloatBarrier
\section{Recommendations}
Authors should be explicit about whether they are developing speech or lipreading systems so that the community can effectively compare methodologies. Where possible, authors are encouraged to make code and data available, and, if not possible, the data should be clearly described such that the approach can be reproduced.

We have clarified the definition of speaker dependent machine lipreading, and authors should carefully consider the split of training, validation and test data prior to model training.


To compare performance we have suggested a simple new notation and we suggest calculating performance on the inverse probabilities (the harder problem). Whilst this in the short-term will reduce accuracy scores, in the long-term it will encourage recognition of unknown or evolving words. We recommend that if the top five values are to be reported, then the top one accuracy should be included separately, alongside the unit \{v, p, or w\}.

\section{Summary}
To conclude our mini review of machine lipreading, we summarise that there is a clear differentiation between machine lipreading and speech reading. It is important that future researchers use the correct terminology in future publications to help the community understand what data conclusions have been drawn upon.
We have also discussed the challenge of speaker independence in lipreading, showing that it can stem from the feature extraction method and the training  parameters of individual speakers.
Furthermore, we have reviewed many of the influences on accuracy scoring in publications from different fields and recommended a new notation to help compare results in the future. 

As a final thought; the fundamental motivation for lipreading is the ability to understand speech when the audio channel is hampered by noise. Therefore is is essential that future work includes acoustically challenging speech environments where the audio channel can not be recognised. It is exciting that there is renewed interest for such a difficult challenge from more research communities. With new ideas and fresh perspectives, we hope that robust, speaker independent machine lipreading in the wild will become a reality. 

\bibliography{egbib}
\end{document}